\documentclass[letterpaper, 10 pt, conference]{ieeeconf}  

\IEEEoverridecommandlockouts                              

\usepackage{graphicx} 
\usepackage{amsmath} 
\usepackage{amsfonts,amssymb}
\usepackage[dvipsnames]{xcolor}
\usepackage{algorithm}
\usepackage{algpseudocode}
\usepackage{tabularx, booktabs, makecell, caption}
\usepackage{siunitx}
\usepackage{afterpage}
\usepackage{flafter}
\usepackage{url}
\usepackage{multirow}

\makeatletter
\def\algbackskip{\hskip-\ALG@thistlm}
\makeatother

\title{\LARGE \bf
RoboNurse-VLA: Robotic Scrub Nurse System based on Vision-Language-Action Model
}
\vspace{0.4cm}
\author{Shunlei Li$^{1}$, Jin Wang$^{2}$, Rui Dai$^{2}$, Wanyu Ma$^{3}$, Wing Yin Ng$^{3}$, Yingbai Hu$^{1}$, and Zheng Li$^{1,3}$
\vspace{0.6cm}
\\\hspace{-0.2cm}
\includegraphics[width=1\textwidth]{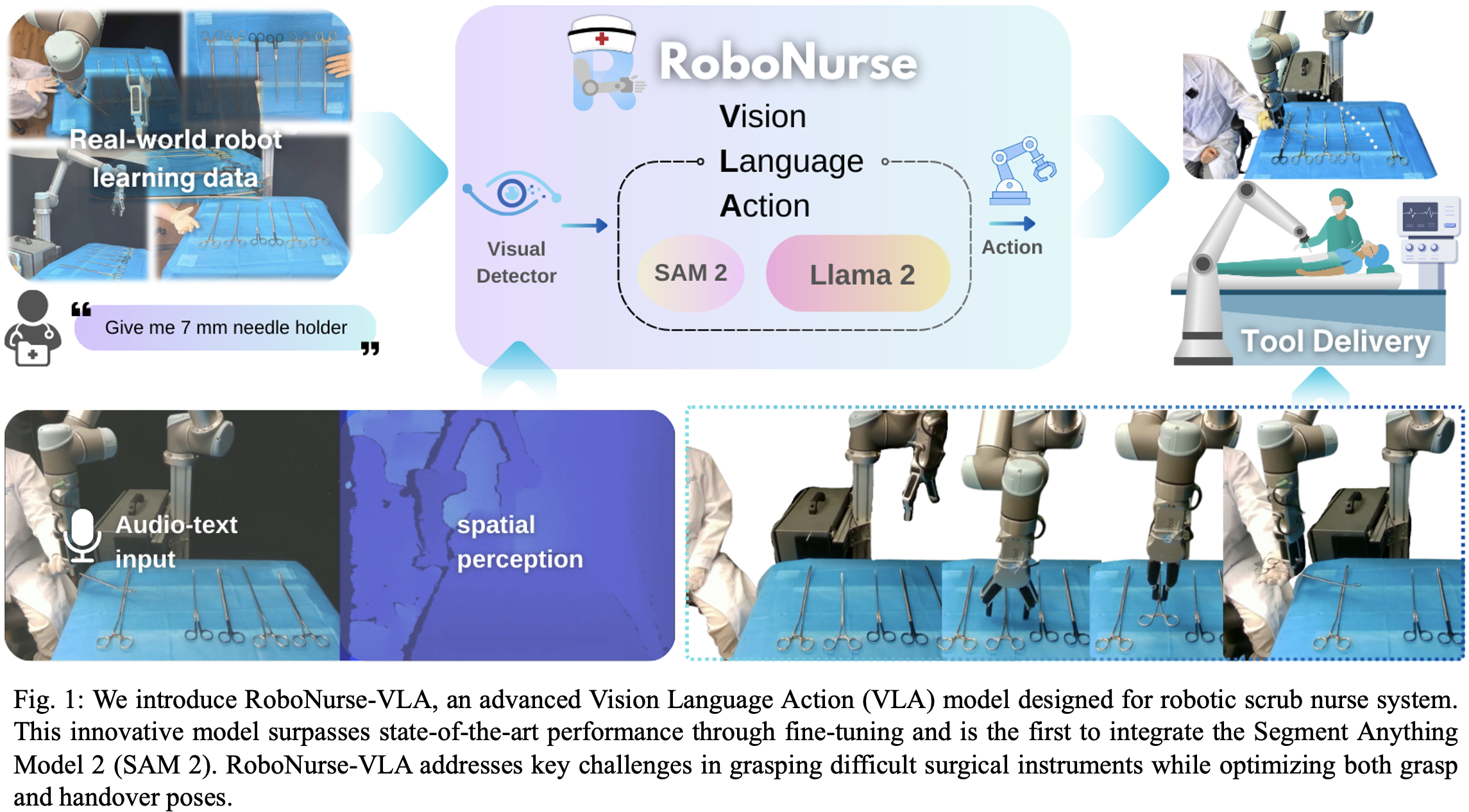}
\vspace{-0.5cm}
\thanks{$^{1}$Multi-Scale Medical Robotics Centre, Ltd., The Chinese University of Hong Kong, Hong Kong, China.
        {\tt\small shunlei.li@outlook.com}}%
\thanks{$^{2}$Humanoids and Human-Centered Mechatronics (HHCM), Istituto Italiano di Tecnologia, Via Morego 30, Genoa, 16163, Italy.}%
\thanks{$^{3}$Department of Surgery, The Chinese University of Hong Kong, Hong Kong, China}%
}

\begin{document}

\maketitle
\thispagestyle{empty}
\pagestyle{empty}


\begin{abstract}
In modern healthcare, the demand for autonomous robotic assistants has grown significantly, particularly in the operating room, where surgical tasks require precision and reliability. 
Robotic scrub nurses have emerged as a promising solution to improve  efficiency and reduce human error during surgery. However, challenges remain in terms of accurately grasping and handing over surgical instruments, especially when dealing with complex or difficult objects in dynamic environments.
In this work, we introduce a novel robotic scrub nurse system, RoboNurse-VLA, built on a Vision-Language-Action (VLA) model by integrating the Segment Anything Model 2 (SAM 2) and the Llama 2 language model.
 The proposed RoboNurse-VLA system enables highly precise grasping and handover of surgical instruments in real-time based on voice commands from the surgeon. Leveraging state-of-the-art vision and language models, the system can address key challenges for object detection, pose optimization, and the handling of complex and difficult-to-grasp instruments. Through extensive evaluations, RoboNurse-VLA demonstrates superior performance compared to existing models, achieving high success rates in surgical instrument handovers, even with unseen tools and challenging items. This work presents a significant step forward in autonomous surgical assistance, showcasing the potential of integrating VLA models for real-world medical applications. More details can be found at \url{https://robonurse-vla.github.io}.
\end{abstract}

\section{Introduction}

\setcounter{figure}{1}
\begin{figure*}
\centerline{\includegraphics[width= 0.99 \linewidth]{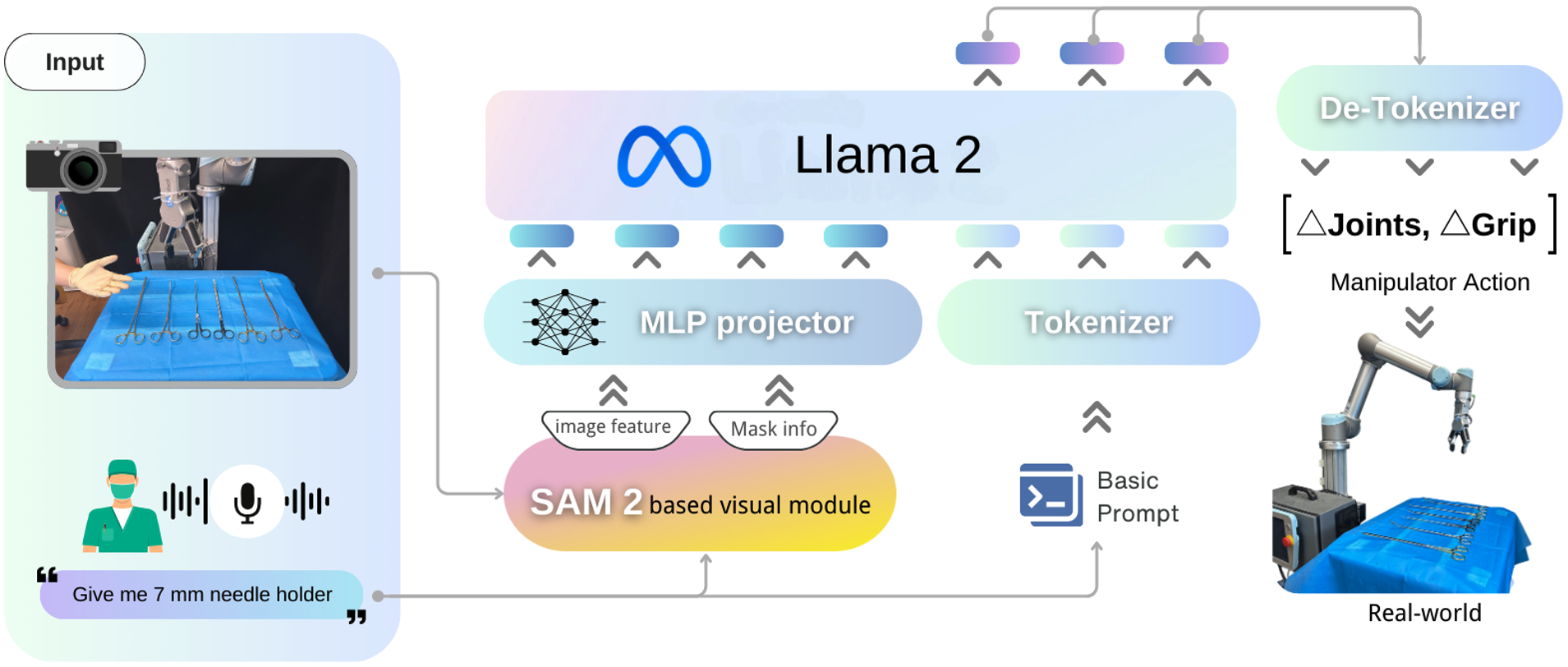}}
\caption{RoboNurse-VLA model architecture. Given an image observation and a speech instruction, the model predicts robot control actions. The architecture consists of three key components: (1) a SAM 2 based vision module, (2) a projector that maps visual features to the language embedding space, and (3) the pretrained Llama 2 7B-parameter LLM in OpenVLA.}
\label{framework}
\vspace{-10pt}
\end{figure*}

A scrub nurse plays a crucial role in the surgical team, assisting the surgeon and other staff to ensure procedures run smoothly \cite{scrubnurse2014review, Korkiakangas2014}. One of the key responsibilities of a scrub nurse is the transfer of surgical instruments. However, this task is labor-intensive and repetitive, leading to increased chances of human error, especially as the scrub nurse becomes fatigued during long surgeries. Additionally, the sharp metal tips of surgical instruments pose a risk of injury to the scrub nurse \cite{watt2008scalpel}. Consequently, the research community has devoted considerable effort to addressing these issues by developing various robotic scrub nurse prototypes to manage the instrument transferring task.

A variety of traditional approaches has been reported for the instrument transfer task.
In \cite{carpintero2010scrub}, Carpintero et al. developed a prototype using contour and template matching for instrument recognition, with a hidden Markov model-based speech recognition system for simple commands. In \cite{jacob2011scrub}, Jacob et al designed a system using fingertip counting as input, with instruments in fixed locations, eliminating the need for a vision module. In  \cite{muralidar2021scrub}, Muralidhar et al. used ArUco markers for instrument localization, allowing requests via speech or keyboard. In \cite{ezzat2021scrub}, Ezzat et al. demonstrated an eye-gaze-controlled robotic scrub nurse, but it only supported fixed instrument positions. Recent efforts have also focused on using deep learning models for ex-vivo instrument detection to assist robotic grasping \cite{akito2022scrub, lavado2018dataset, mark2022dataset, 9426258, 9863381}.
However, the works mentioned above primarily focused on the 'instrument recognition method' and 'user input method,' without adequately addressing the subsequent step of surgical instrument transfer, known as the 'handover.'

Several learning-based approaches have been explored to address the 'handover' process, however, the learned policies for robotic manipulation often struggle with new environments and unfamiliar tasks, lacking robustness beyond their training scenarios \cite{cobbe2019Quantifying, chi2024diffusionpolicyvisuomotorpolicy, xie2023decomposinggeneralizationgapimitation}. While foundation models like CLIP \cite{radford2021learningtransferablevisualmodels}, SigLIP \cite{zhai2023sigmoidlosslanguageimage}, and Llama 2 \cite{touvron2023llama2openfoundation} outperform in generalization due to the priors acquired from their extensive Internet-scale pretraining datasets, their direct replication in the application of robotic manipulation remains underexplored due to limited datasets, such as Octo \cite{octomodelteam2024octoopensourcegeneralistrobot}.
Recent works have investigated the combination of pretrained language and vision-language models for robotic representation learning \cite{nair2022r3muniversalvisualrepresentation, karamcheti2023languagedrivenrepresentationlearningrobotics} and task planning \cite{stone2023open}\cite{wang2024autonomousbehaviorplanninghumanoid}\cite{wang2024hypermotion}. 
These models can be leveraged to create vision-language-action models (VLAs) for controlling robots, with visually-conditioned language models (VLMs) being refined to generate robot control actions.
However, these VLAs did not show their ability in surgical handover task with structurally complex and similar objects.

In this paper, we propose a framework for robotic scrub nurse system by using advanced VLA integrating Segment Anything Model 2 (SAM 2) \cite{ravi2024sam2} and Llama 2 \cite{touvron2023llama2openfoundation}, as shown in Fig.~1.
It leverages the semantic understanding capabilities of Large Language Model (LLM) in reaction to human instructions, which is based on the knowledge of the surgical environment and the manipulation actions possessed by the robot. When a doctor provides the instruction, the VLA will detect the corresponding instrument and generate handover actions to deliver the tools to the doctor.
This approach can generate autonomous surgical handover actions based on the environment of surgical instruments and the surgeon's hands.

The main contributions can be summarized as follows:
\begin{itemize}
\item We developed an innovative VLA model that allows fine-tuning and surpasses the performance of state-of-the-art models, which is the first application to surgical instrument handover tasks.
\item We designed a SAM 2 based vision module and integrated it with VLA models, enabling language-guided grasping of challenging items. 
\item Our method effectively addresses challenges related to distinguishing similar items, as well as optimizing both the grasp and handover poses. 
\end{itemize}

We validated the feasibility of RoboNurse-VLA through real-world experiments and highlighted its efficiency in handling various tasks by comparing with mutiple baselines.

\section{Related works}
VLMs use large-scale image and language data to generate natural language, enabling applications like visual question answering \cite{hudson2019gqa,singh2019towards,bai2024surgical} and various visual recognition tasks \cite{zhang2024vision}. 
These models connect pretrained vision encoders with language models, leveraging advancements in computer vision and language processing for multimodal tasks. 
Initially, cross-attention methods were explored \cite{li2023blip, laurenccon2024obelics}, but newer open-source VLMs now use a simpler ``patch-as-token'' method, treating visual patches as tokens for language models \cite{karamcheti2024prismatic, kim2024openvla}. 

Several works have investigated using VLMs in robotics by integrating them into end-to-end visuomotor manipulation policies \cite{stone2023open}. However, these approaches often introduce significant structural complexity or require calibrated cameras, limiting their broader applicability. More recent efforts have followed similar approaches to ours by fine-tuning large pretrained VLMs to predict robot actions \cite{padalkar2023open, li2023vision}. These models, commonly known as VLAs, merge robot control actions directly into the VLM architecture.

Most closely related, RT-2-X \cite{padalkar2023open}, a 55B-parameter VLA policy trained on the Open X-Embodiment dataset, achieves state-of-the-art performance in generalist manipulation tasks. Similarly, OpenVLA \cite{kim2024openvla}, a 7B-parameter open-source VLA model trained on 970k real-world robot demonstrations, delivers robust cross-embodiment robot control straight out of the box. 

Our work diverges from RT-2-X and OpenVLA in several key areas: by integrating a robust VLM with our robotic scrub nurse dataset, our approach surpasses both RT-2-X and OpenVLA in performance, despite being significantly smaller in scale; and we are the first to integrate SAM 2 \cite{ravi2024sam2}, a state-of-the-art foundation model towards solving promptable visual segmentation in images and videos, into VLAs to enable the grasping of difficult-to-hold items, such as thin sticks and balls.

\section{RoboNurse-VLA Model}
In this section, we introduce the RoboNurse-VLA model, which enables surgeons to voice-control a robot to grab targeted items and place them in the surgeon's hand. 
Specifically, when the surgeon names the desired surgical instrument and extends their hand, the robot can accurately identify and grasp the correct item, then present it to the surgeon in an optimal position for easy retrieval.
The model is adaptable to new robots and tasks of difficult-to-hold items, accommodating different sensory inputs and action spaces through fine-tuning.

\subsection{Architecture}

\begin{figure}
\centerline{\includegraphics[width=0.99 \linewidth]{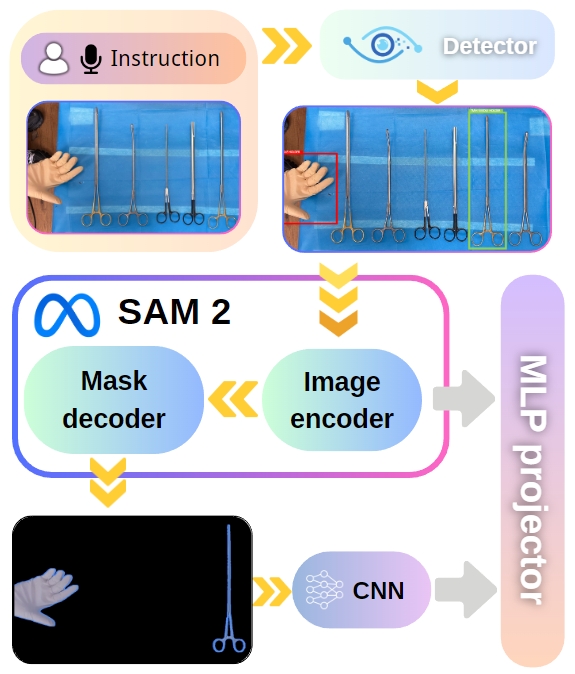}}
\caption{The workflow of vision module.}
\label{vision}
\end{figure}

The architecture of recent VLMs typically includes three key components: (1) a visual encoder that transforms image inputs into a set of “image patch embeddings”; (2) a projector that converts these embeddings into a format compatible with a language model; and (3) a large language model (LLM) as the core backbone \cite{karamcheti2024prismatic, kim2024openvla}. During training, the VLM is optimized end-to-end using a next text token prediction task on vision and language data that is paired or interleaved, sourced from various Internet datasets.

RoboNurse-VLA follows the same standard architecture described above, with a 600M-parameter visual module, a small 2-layer Multi-Layer Perceptron (MLP) projector, and a pretrained 7B-parameter Llama 2 language model backbone of OpenVLA. In addition, to enhance user convenience, the model also includes a deep learning-based Gladia Automatic Speech Recognition (ASR).

The framework of RobotNurse-VLA is illustrated in Fig. \ref{framework}. The input comprises visual data from a depth camera and voice commands from the operator. Once the speech is transcribed into text, it serves as prompts for Llama 2. In parallel, the vision module uses the text to detect the target, producing both global image features and segmentation masks for the instrument and hand. These image features and masks are processed through a MLP and provided as input to Llama 2. Llama 2 then generates action commands, which are passed through a De-Tokenizer and sent to the robotic arm for task execution.

Notably, in the visual module as shown in Fig. \ref{vision}, a pretrained custom detector is utilized to identify both the target and the hand, providing bounding boxes to the image encoder of SAM 2. Specifically, we use a pretrained YOLOv8 model tailored for surgical instruments. This detector can be replaced or customized according to user requirements.
The mask decoder of SAM 2 then segments both identified objects.
The features from the SAM 2 image encoder and the mask produced by the mask decoder are projected into Llama 2 as patches through an MLP, where the resulting feature vectors are concatenated channel-wise. 
Before concatenating with image features, convolutional layers are employed to preserve the spatial information in the mask. 
The network consists of three Conv2D layers, followed by Global Average Pooling and a Fully Connected Layer, ensuring that both the spatial relationships and key features of masks are retained throughout the process.
The image encoder enables the model to observe the entire environment, while the segmented masks allow the model to focus specifically on the target and the surgeon's hand position.

\subsection{Data collection}

\begin{figure}
\centerline{\includegraphics[width=8cm]{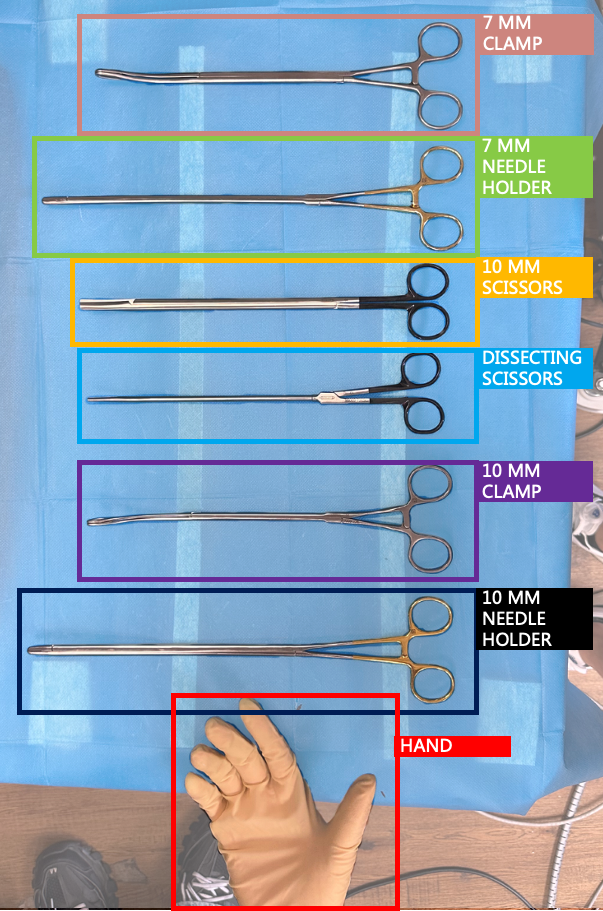}}
\caption{The category of ex-vivo surgical instruments and hand.}
\label{item}
\end{figure}

To support research in ex-vivo instrument and surgeon's hand detection of YOLOv8, we have created a surgical dataset. This dataset includes images of the hand and six distinct but visually similar instruments as shown in Fig. \ref{item}, captured in various combinations, orientations, and poses. It comprises 700 images in total, all taken with an RGB-DSLR camera at a consistent resolution of 6000x4000 pixels, with each class represented by 100 instances. The images were manually annotated using LabelImg, an open-source annotation tool \cite{labelImg2015}.

\begin{figure}
\centerline{\includegraphics[width=8cm]{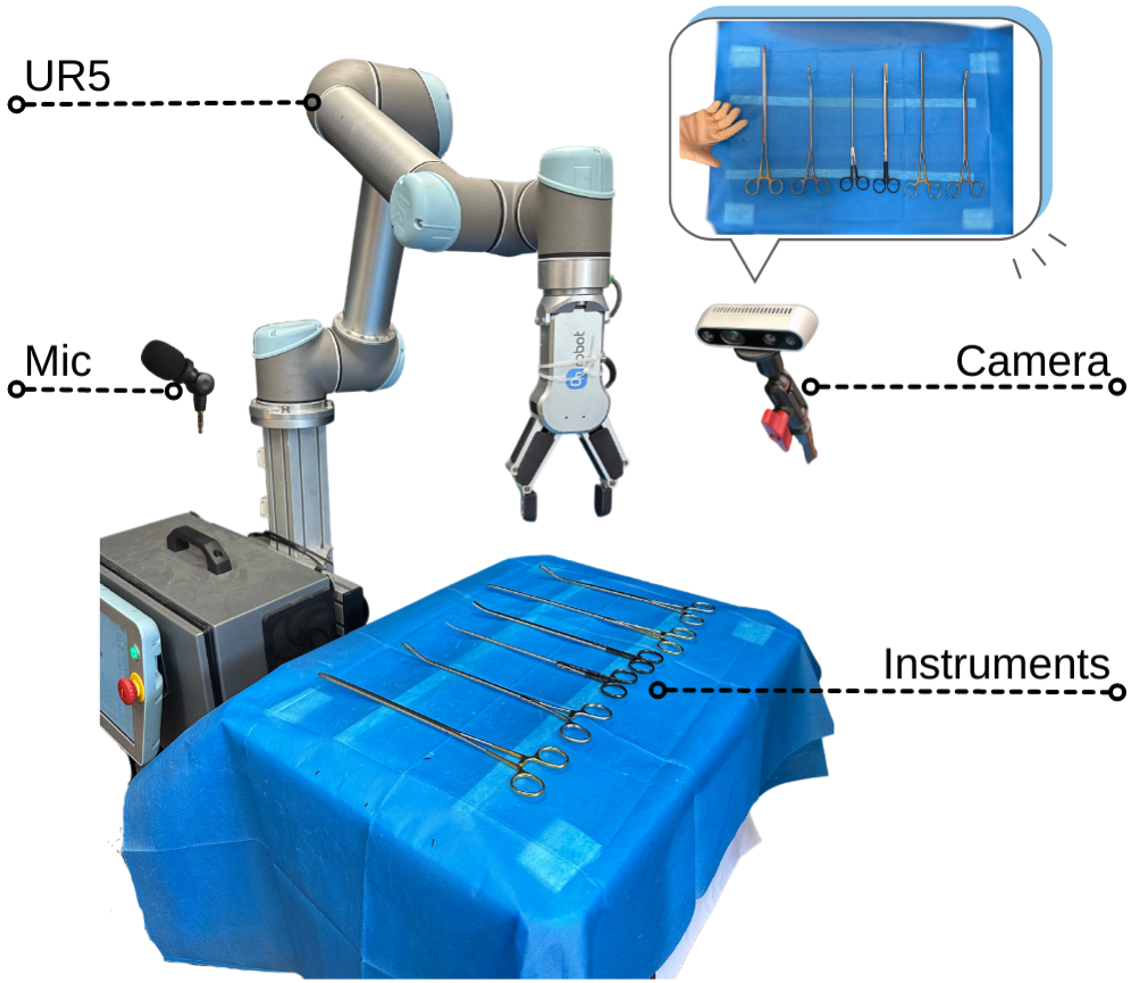}}
\caption{The setup of RoboNurse-VLA includes a microphone, a table with six surgical instruments, an Intel RealSense Depth Camera D415, and a UR5 robotic arm.}
\label{setup}
\end{figure}

The system consists of a microphone, a table with six surgical instruments, an Intel RealSense Depth Camera D415, a UR5 robotic arm, and a workstation for real-time inference as shown in Fig. \ref{setup}. A simple joint position control system operates at 30 Hz, where actions are defined as joint targets and converted into motor control signals using a high-frequency PD controller. Additionally, we offer an end-effector controller where actions are specified by both the position and orientation of the gripper. 
All data and results presented in this paper were obtained using the joint position controller. 

We frame the task as a "vision-language" problem, where an input image and a speed voice task instruction is recognized and mapped to predicted robot actions. To integrate robot actions into the language model's output, we discretize continuous actions into 256 bins per dimension and represent them as discrete tokens \cite{brohan2023rt}. Due to limitations in the Llama tokenizer, we overwrite the 256 least-used tokens with our action tokens. RoboNurse-VLA is then trained using a next-token prediction objective, focusing on the action tokens. 

We developed our data collection protocol to facilitate learning for the scrub nurse handover task. To enhance task robustness, data was collected across diverse environments, incorporating intentional variations in object placement, camera angles, hand postures (including height changes), and workspace layouts. The dataset comprises 648 demonstrations generated from predefined scripted policies, with each entry containing both trajectory data and corresponding images.

\subsection{Training}

The resolution of input images significantly influences the computational demands during VLA training. Higher-resolution images produce more patch tokens, resulting in longer context lengths, which leads to a quadratic increase in computational requirements. To balance this, we chose a resolution of 224 × 224 pixels for the final RoboNurse-VLA model.

For training, we froze the vision module and fine-tuned the pretrained LLaMA 2 model from OpenVLA for predicting robotic actions. We used LoRA (Low-Rank Adaptation), a fine-tune method, which applies multiple rank values across all linear layers of the model \cite{hu2021lora}.

With this setup, we successfully fine-tuned RoboNurse-VLA for the robotic scrub nurse task in just 20 hours using a single A100 GPU.

\section{Experiments and results}
The objective of our experimental evaluation is to assess RoboNurse-VLA’s effectiveness as a robust, out-of-the-box control policy for robotic scrub nurse tasks. Specifically, the experiments include:

\begin{itemize} 
\item Evaluating the accuracy of Gladia ASR, the pretrained detector, and SAM 2. 
\item Testing the zero-shot performance of Octo, RT-2-X, OpenVLA, and RoboNurse-VLA across various tasks. 
\item Assessing the results of fine-tuned models for the scrub nurse handover tasks. 
\item Evaluating the performance of fine-tuned models on unseen tools and challenging-to-grasp items. 
\end{itemize}

\subsection{Accuracy of Gladia ASR, Detector, and SAM 2}
To ensure the correctness of RoboNurse-VLA, we first tested the accuracy of Gladia ASR, the detector and SAM 2.
Gladia ASR achieves an accuracy of 99.5$\%$ for commands related to surgical instruments. The pretrained YOLOv8 attains an average precision of 99.2$\%$ and an average recall of 99.4$\%$ on our dataset. SAM 2 achieves a Dice score of 0.962, an IoU of 0.933, and an MAE of 0.009 for segmenting instruments in our dataset, using the bounding boxes generated by YOLOv8.

\subsection{Zero-shot performance}

\begin{figure}
\centerline{\includegraphics[width=8cm]{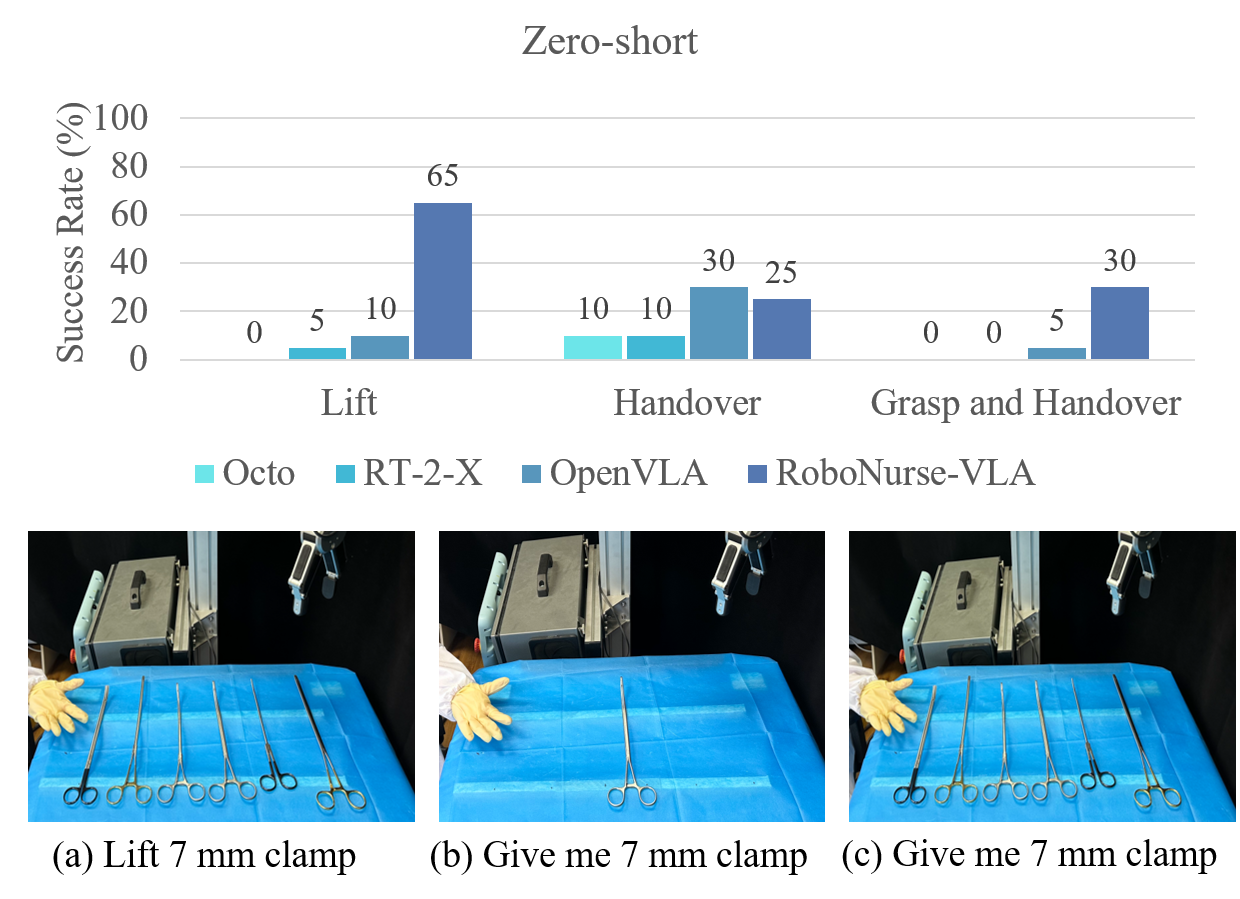}}
\caption{Zero-short evaluation tasks and results.}
\label{zeroshort}
\end{figure}

\begin{figure}
\centerline{\includegraphics[width=8cm]{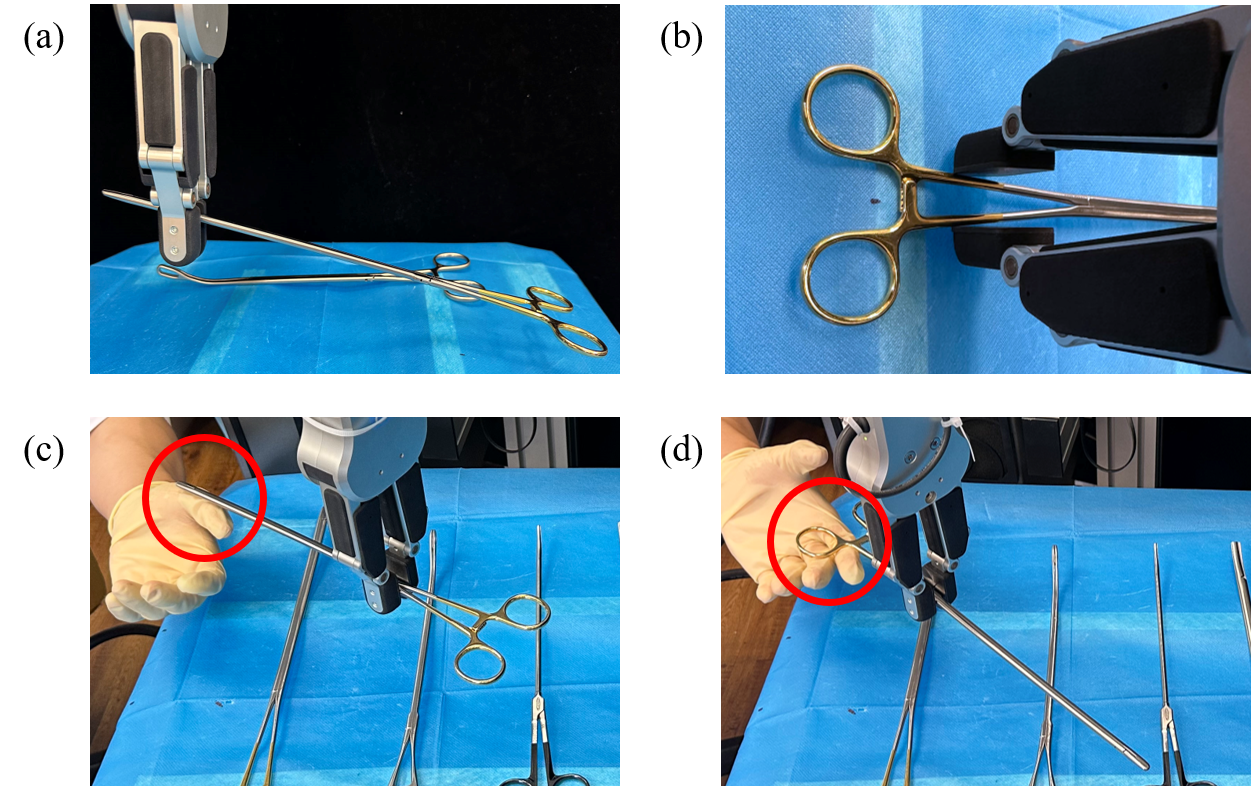}}
\caption{The examples for incorrect grasping and handover.}
\label{wrong}
\end{figure}

In this experiment, RoboNurse-VLA was not trained on the scrub nurse dataset, as the goal is to evaluate the functionality of our visual module and compare it with Octo, RT-2-X, and OpenVLA. We conducted three tasks with 20 trials each, as shown in Fig. \ref{zeroshort}: (a) Six visually similar surgical instruments are arranged, and the instruction is `lift \{instrument\}', such as a 7mm clamp; (b) A single instrument is placed, and the command is `give me \{instrument\}'; (c) Six visually similar surgical instruments are arranged again, and the command is `give me \{instrument\}'.

The results are summarized in Fig. \ref{zeroshort}.
In task (a), we observed that Octo, RT-2-X, and OpenVLA struggled with the tested tasks, often failing to manipulate the correct object. This was due to the instruments being unseen in their training data and the difficulty in identifying the correct one. 
Thanks to the powerful pretrained detector and SAM 2, Robonurse-VLA was able to correctly identify the target for grasping. However, since our model has not yet learned how to properly grasp surgical instruments, some failures occurred as shown in Fig. \ref{wrong} (a) and (b), resulting in a success rate of 65\%.
In the second task, although only a single instrument was involved, the success rate did not improve significantly due to incorrect grasping postures. Additionally, many instances could not be considered successful during handover, as shown in Fig. \ref{wrong} (c) and (d). In this study, a successful handover is defined as the ability to grasp the instrument when the hand performs a clasping motion at its current position.
In the final task, we evaluated the ability of four models to identify, grasp, and hand over instruments in a typical scenario. The overall results were unsatisfactory, with only RoboNurse-VLA achieving a success rate of 30\%.

\begin{table*}[t]
\centering
\caption{Performance comparison of RoboNurse-VLA with other models for scrub nurse tasks.}
\begin{tabular}{c c c c c c c}
\hline\hline 
\multicolumn{2}{c}{\textbf{Experiment \& Task}} & \textbf{Octo} & \textbf{RT-2-X} & \textbf{Diffusion Policy} & \textbf{OpenVLA} & \textbf{RoboNurse-VLA} \\
\hline
\multirow{3}{*}{Zero-shot} 
 & Lift & 0\% & 5\% & N/A & 10\% & \textbf{65\%} \\
 & Handover (single) & 10\% & 10\% & N/A & 30\% & \textbf{25\%} \\
 & Handover (multiple) & 0\% & 0\% & N/A & 5\% & \textbf{30\%} \\
\hline
\multirow{3}{*}{Fine-tuned} 
 & On table & 20\% & N/A & 15\% & 45\% & \textbf{100\%} \\
 & Height change & 10\% & N/A & 10\% & 40\% & \textbf{95\%} \\
 & Pose change & 0\% & N/A & 0\% & 25\% & \textbf{95\%} \\
\hline
\multirow{2}{*}{Fine-tuned} 
 & Unseen tool & 15\% & N/A & 15\% & 35\% & \textbf{90\%} \\
 & Difficult-to-grasp & 20\% & N/A & 25\% & 40\% & \textbf{95\%} \\
\hline\hline 
\end{tabular}
\label{tab.Performance_results}
\end{table*}

\subsection{Fine-tuned models for scrub nurse handover}

\begin{figure}
\centerline{\includegraphics[width=8cm]{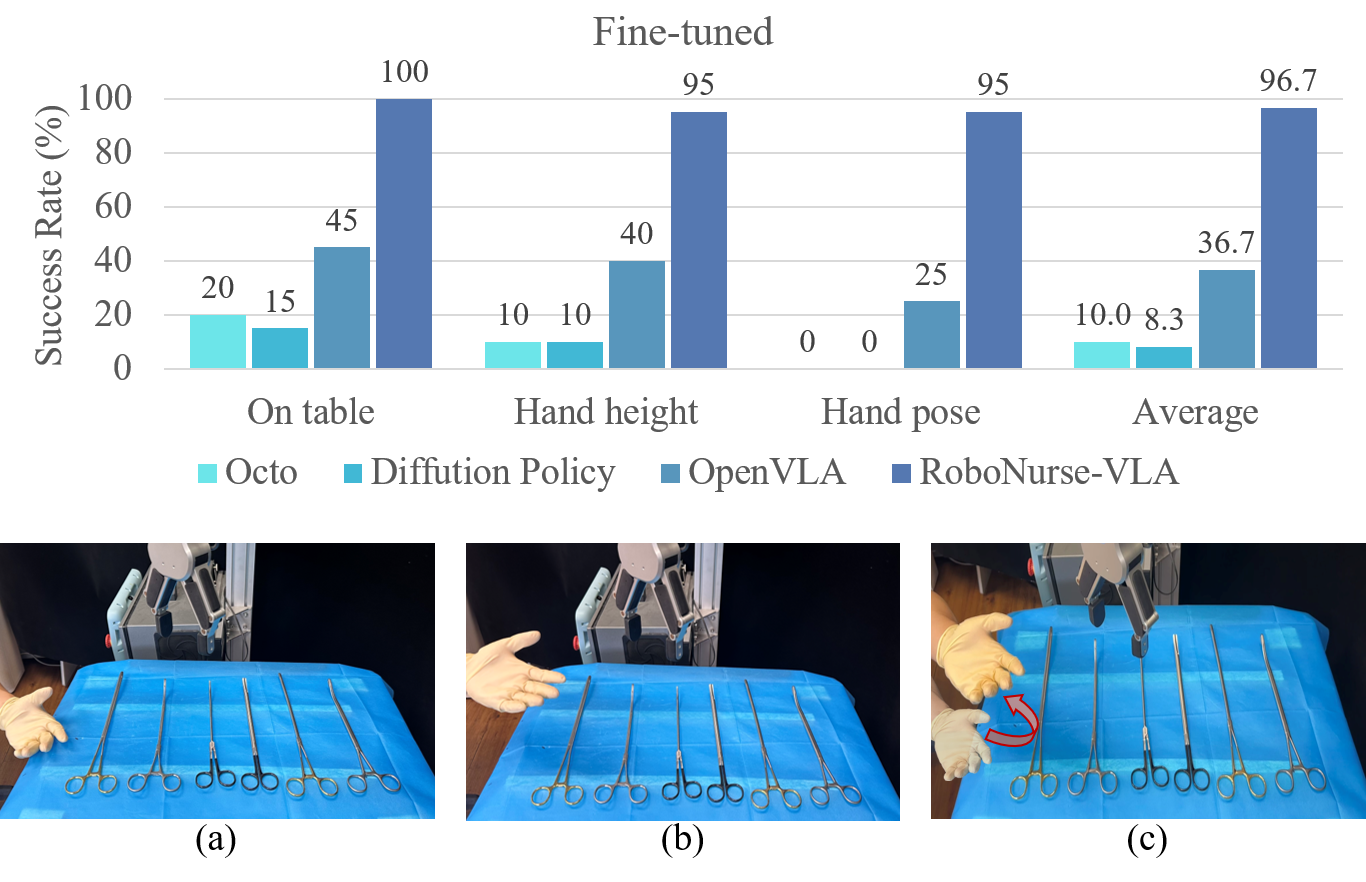}}
\caption{The evaluation tasks and results of fine-tuned models.}
\label{finetune}
\end{figure}

\begin{figure}
\centerline{\includegraphics[width=8cm]{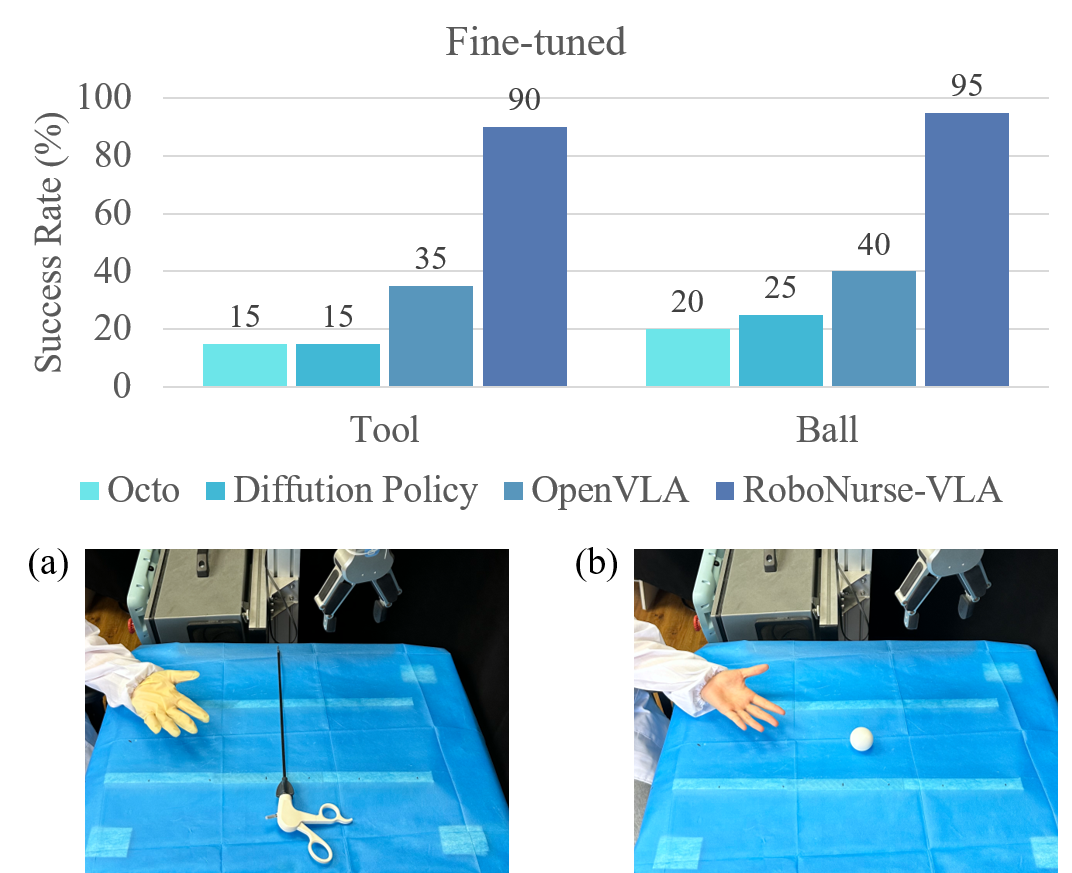}}
\caption{The evaluation tasks and results for unseen tools and difficult-to-grasp items.}
\label{hard}
\vspace{-10pt}
\end{figure}

In this experiment, we aim to evaluate the performance of RoboNurse-VLA and compare it with Octo, OpenVLA, and Diffusion Policy. Since fine-tuning of RT-2-X is not supported through its inference API, we compare it with the Diffusion Policy, which aligns with the input and output specifications of the experiment.

The tasks are illustrated in Fig. \ref{finetune}: (a) the scenario involves a hand and instruments placed on the table; (b) the hand position varies in height across different trials; (c) the hand position changes during a single trial. The instruction given is to `give me \{instrument\}'.

RoboNurse-VLA achieved a 100\% success rate, whereas other models struggled with task (a). Our method precisely detects the object, grasps it correctly, and delivers it to the hand in a comfortable pose. Although the success rates of the other methods improved after fine-tuning, errors like those shown in Fig. \ref{wrong} remained frequent.

In the second task, both OpenVLA and RoboNurse-VLA showed only slight decreases in performance, demonstrating robustness to changes in spatial positioning. Octo and Diffusion Policy already exhibited low performance, so further accuracy degradation was not meaningful.

For the final task, Octo and Diffusion Policy failed to complete the task. While OpenVLA handled hand pose changes during task execution to some extent, achieving a success rate of 25\%, RoboNurse-VLA maintained a high accuracy of 95\% with acceptable latency of 5 hz inference, unaffected by these changes. On average, our method significantly outperforms other state-of-the-art models in the scrub nurse handover scenario.

\subsection{Performance on unseen tools and difficult-to-grasp items}

In the final experiment, we first evaluate the performance on unseen instruments, followed by testing the ability to handle difficult-to-grasp items, such as a ping-pong ball. The scenarios are illustrated in Fig. \ref{hard}: (a) and (b) show that both the object and the hand are placed on the table with no change in pose during the trial.

Since the unseen instruments have similar shapes to those in our dataset, the fine-tuned models exhibit a performance close to that in the previous experiment, with only a slight decrease. RoboNurse-VLA still achieves the highest success rate at 90

For difficult-to-grasp items, like a ping-pong ball, we had to set an appropriate gripping force. If the force exceeds 4N, the ping-pong ball will break, so we set the gripper force to 4N. Additionally, the gripper needs to grasp the center of mass; otherwise, the ball will slip or bounce away.
Ultimately, our method achieved a 95\% success rate, while the success rates of the other models remained below 50\%.

The performance results are presented in Table \ref{tab.Performance_results}, where RoboNurse-VLA consistently outperforms all other models across the evaluated experiments.

\section{Conclusion}
In this paper, We proposed an innovative RoboNurse-VLA system that combines cutting-edge vision and language models to automate the role of a robotic scrub nurse with inference of 5 hz. By integrating SAM 2 and Llma 2, RoboNurse-VLA effectively addresses challenges related to surgical handover tasks, such as precise object detection, grasp optimization, and handling of difficult-to-grasp tools. Our experimental results show that RoboNurse-VLA significantly outperforms existing state-of-the-art models like Octo, OpenVLA, and Diffusion Policy, achieving high success rates even with unseen and complex objects. The system's adaptability and robust performance highlight its potential for enhancing surgical efficiency and safety.

Future work will focus on adapting RoboNurse-VLA for real clinical application, including expanding the dataset and testing in complex clinical environments. We aim to address safety concerns by developing obstacle avoidance techniques and improving human-robot interactions to ensure seamless collaboration with surgical teams. Additionally, real-world trials will be conducted to validate the system’s performance in diverse and dynamic settings.







\bibliographystyle{IEEEtran}
\bibliography{references}

\end{document}